\documentclass{bmvc2k}
\usepackage{float}
\usepackage{setspace}
\definecolor{ocre}{RGB}{25,25,112}
\usepackage{caption}
\usepackage[font={color=ocre},figurename=Figure,labelfont={it}]{caption}
\usepackage{multirow}
\usepackage{amsfonts}
\usepackage{tipa}
\usepackage[normalem]{ulem} % use normalem to protect \emph %DIF > 
\usepackage{bm}
\newcommand\hl{\bgroup\markoverwith {\textcolor{yellow}{\rule[-.5ex]{2pt}{2.5ex}}}\ULon}
\title{Synchronous Bidirectional Learning for Multilingual Lip Reading}

\addauthor{Mingshuang Luo}{mingshuang.luo@vipl.ict.ac.cn}{1,2}
\addauthor{Shuang Yang}{shuang.yang@ict.ac.cn}{1}
\addauthor{Xilin Chen}{xlchen@ict.ac.cn}{1,2}
\addauthor{Zitao Liu}{liuzitao@100tal.com}{3}
\addauthor{Shiguang Shan}{sgshan@ict.ac.cn}{1,2,4}
\addinstitution{
 Key Lab of Intelligent Information
Processing of Chinese Academy of
Sciences(CAS), Inst. of Computing \\Technology, CAS,
Beijing, China
}
\addinstitution{
University of Chinese Academy of\\
Sciences,
Beijing, China
}
\addinstitution{
TAL Education Group,
Beijing, China
}
\addinstitution{CAS Center for Excellence in Brain Science 
and Intelligence Technology,
Shanghai, China}

\runninghead{Luo, Yang, Chen, Liu, Shan}{SBL for Multilingual Lip Reading}

\begin{document}

\maketitle

\begin{abstract}
	% 回答两个问题：
	% (1) 多语言的lip reading是否可行：分类的各种组合、解码的各种组合
	% (2) 提出一种同步双向解码的多语言lip reading
	% (3) 达到sota
	% 
% 本文关注对象：多语言联合的lip reading
% 出发点：每种语言的音素之间相互连接是有一定规律的；如果模型掌握了不同的语言所对应的这种规律，那它就可以做多语言的lip reading了
% 
% 基于此，我们提出了SBL框架。将语言规律的学习问题，转化为给定上下文的情况下对当前单元的推断问题。
% 
Lip reading has received increasing attention in recent years. This paper focuses on the synergy of multilingual lip reading. There are about as many as 7000 languages in the world, which implies that it is impractical to train separate lip reading models with large-scale data for each language. Although each language has its own linguistic and pronunciation rules, the lip movements of all languages share similar patterns due to the common structures of human organs. Based on this idea,  we try to explore the synergized learning of multilingual lip reading in this paper, and further propose a \textit{synchronous bidirectional learning} (SBL) framework for effective synergy of multilingual lip reading. 
We firstly introduce phonemes as our modeling units for the multilingual setting here. Phonemes are more closely related with the lip movements than the alphabet letters. At the same time, similar phonemes always lead to similar visual patterns no matter which type the target language is. Then, a novel SBL block is proposed to learn the rules for each language in a fill-in-the-blank way. Specifically, the model has to learn to infer the target unit given its bidirectional context, which could represent the composition rules of phonemes for each language. To make the learning process more targeted at each particular language, an extra task of predicting the language identity is introduced in the learning process. Finally, a thorough comparison on LRW (English) and LRW-1000 (Mandarin) is performed, which shows the promising benefits from the synergized learning of different languages and also reports a new state-of-the-art result on both datasets.
%\vspace{-0.2cm}
\end{abstract}
\vspace{-0.4cm}
\section{Introduction}
\vspace{-0.2cm}
Lip reading aims to infer the speech content by using visual information like lip movements, and is robust to the ubiquitous acoustic noises \cite{luo2020pseudo-convolutional} in our life. 
This special property makes it important for automatic speech recognition in noisy or silent scenarios \cite{chung2017lip, zhang2019spatio-temporal, StafylakisT17, zhang2020can}.
With the rapid development of deep learning technologies and the recent emergence of several large-scale lip reading datasets \cite{chung2016lip, yang2019lrw-1000, chung2017lip, zhao2019cascade}, there have been several appealing results in recent years \cite{assael2017lipnet,afouras2019deep, zhang2019spatio-temporal, zhao2019cascade, zhao2019hearing}.
However, almost all of the existing methods focus on the problem of monolingual lip reading.
In this paper, we try to make an exploration of multilingual lip reading, which has not been considered before to the best of our knowledge.
%\hl{in Fig 1.(c): it is more proper to change ``each pair has'' $\rightarrow$ ``each pair contains''.}
\begin{figure}
    \setlength{\abovecaptionskip}{0.2cm}
    \setlength{\belowcaptionskip}{-0.2cm} 
    \centering
    \includegraphics[width=1.01\textwidth]{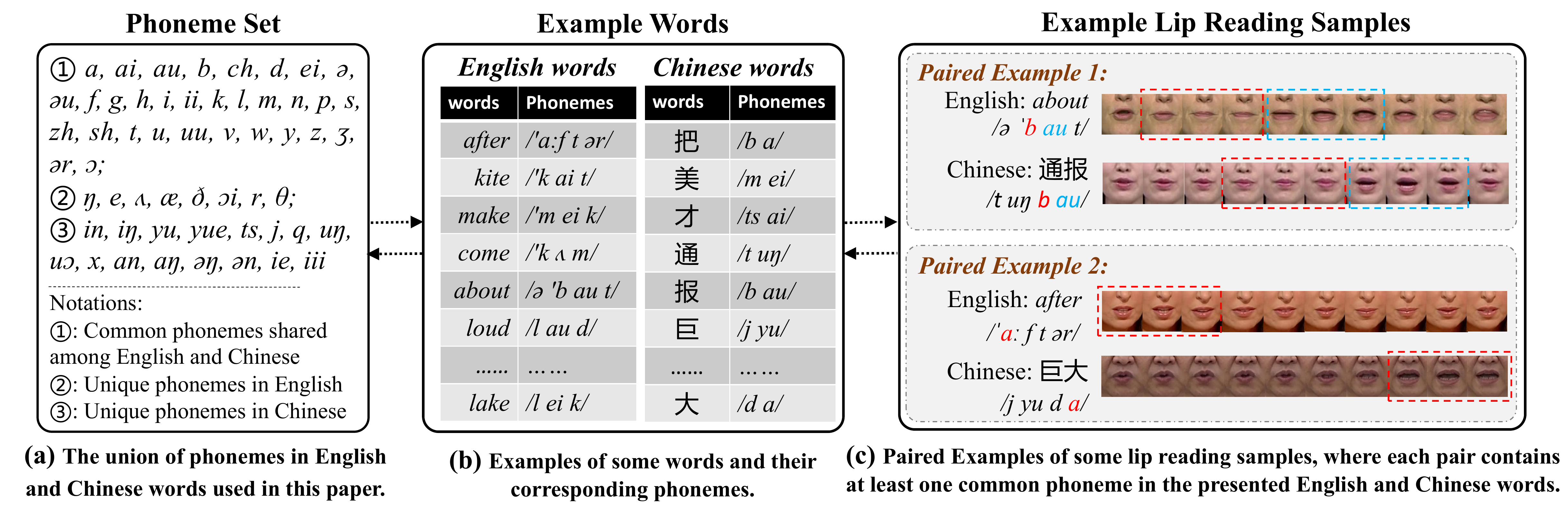}
    \caption{Illustration of the modeling units in our work.}\label{illustration_our_idea}
    \vspace{-0.4cm}
\end{figure}

\vspace{-0.4cm}
Limited by the structure of our vocal organs, the number of distinguishable pronunciations we could make is finite. So the set of distinguishable pronunciations in each language is finite, leading to many common pronunciations shared among different languages.
For example, there are as many as 32 phonemes existing in both English and Mandarin words, as shown in Figure \ref{illustration_our_idea}.(a). Figure \ref{illustration_our_idea}.(b) provide some example words with their corresponding phoneme-based representations. The same phonemes in different languages would generate the same or similar lip movements even though the speakers are of different languages, as shown by Figure \ref{illustration_our_idea}.(c).
Besides, knowledge sharing and transfer among different languages could further help the unique model shared by different languages learn more easily than learning separately from every single language.
These factors make us think it possible to perform a synergize learning of  multilingual lip reading.

Each language has its own rule to compose different units (characters or phonemes) into a valid word. If we could make the lip reading model master the composition rules for each language, it should be able to obtain good recognition results when meeting these languages.
Based on this idea, we consider the learning process of the composition rule for each language as to learn a fill-in-the-blank problem according to the correct rules. 
If the model could make correct predictions for any missing units, no matter which language the input is, as long as its previous and later context is given, then the decoder module should be also effective to compose correct phonemes into correct words in the multilingual lip reading setting.
Therefore, a novel synchronous bidirectional learning (SBL) block is introduced to construct the decoder module to finish our prediction process for the multilingual lip reading problem.

Overall, the main contributions could be summarized as follows.
\vspace{-0.25cm}
\begin{itemize}\setlength{\itemsep}{0cm} 
    \item We make a first exploration to the problem of multilingual lip reading. As far as we know, it is the first time to tackle the lip reading problem in a multilingual setting with large-scale lip reading datasets. 
    \vspace{-0.1cm}
    \item To perform a better multilingual lip reading, we introduce phonemes as the modeling units, which acts as the bridge to link different languages. Then, a novel synchronous bidirectional learning (SBL) framework is proposed to learn the composition rule for each language. Finally, an extra task of judging the language type is introduced to make the learning more targeted at each specific language at present.
    \vspace{-0.1cm}
    \item With a thorough evaluation and comparison, our method not only shows a clear advantage of multilingual lip reading over monolingual lip reading, but also outperforms the existing state of the art performance by a large margin on the benchmarks of different languages.
    \vspace{-0.4cm}
\end{itemize} 

\section{Related work}
\vspace{-0.3cm}
\subsection{Lip Reading}
\vspace{-0.1cm}
Great strides have been made in lip reading recently \cite{zhang2019spatio-temporal, zhang2020can,zhao2020mutual,luo2020pseudo-convolutional, afouras2019deep, chung2017lip, chung2016lip, petridis2018end, zhao2019cascade, zhao2019hearing, StafylakisT17, assael2017lipnet, wang2019multi-grained, xiao2020deformation}. Existing lip reading methods could be generally divided into two categories, decoding based methods and classification based methods. 

In the first category, lip reading is considered as a sequence (image sequence) to sequence (text sequence) problem, and seq2seq models based on RNN or Transformer \cite{vaswani2017attention} are applied. For example, Chung et al. \cite{chung2017lip} was the first to use an RNN based encoder-decoder framework to perform lip reading and has achieved an appealing result. 
Luo et al. \cite{luo2020pseudo-convolutional} proposed to introduce the CER (character error rate) to the RNN based seq2seq model to perform a more direct optimization over the evaluation metric. 
Zhang et al. \cite{zhang2019spatio-temporal} introduced a temporal focal block to capture the short-range dependencies based on the Transformer-seq2seq model. 
%\hl{II} % YS:这里放到实验里讲→：They achieved an accuracy of $83.7$\% on the English benchmark LRW after pre-training on LRS2-BBC and LRS3-TED. It is the current best performance using decoding based methods.

In the second category, the whole input image sequence is taken as a single object belonging to a word class, and the lip reading problem is considered as a video classification problem.
In 2017, Stafylakis et al. \cite{StafylakisT17} proposed an effective pipeline to perform classification based lip reading, which has been used widely in the subsequent lip reading methods \cite{petridis2018end, zhang2019spatio-temporal, luo2020pseudo-convolutional, xiao2020deformation, zhao2020mutual, zhang2020can}.
Later, Wang \cite{wang2019multi-grained} proposed a multi-grained spatio-temporal model to perform lip reading by collecting information from three different granularities.  
% 
% Zhang et al. \cite{} performed a comprehensive study to evaluate the effects of different facial regions and obtained an accuracy of 85.02$\%$ and 45.24$\%$ on LRW and LRW-1000 respectively. This performance is the current best performance on these two datasets. 
\vspace{-0.2cm}
\subsection{Multilingual Learning}
\vspace{-0.1cm}
Multilingual learning has been studied for a long time in the field of speech recognition and natural language processing. Dalmia et al. \cite{dalmia2018sequence-based} found that an end-to-end multi-lingual training of seq2seq models is beneficial to low resource cross-lingual speech recognition. In 2018, Zhou et al. \cite{zhou2018multilingual} proposed to use the sub-words as modeling units with the Transformer architecture \cite{vaswani2017attention} and achieved good results for multilingual speech recognition. Toshniwal et al. \cite{Toshniwal-multilingual-asr} take a union of language-specific grapheme sets and train a grapheme-based sequence-to-sequence model on data combined by different languages for speech recognition. Besides the design of modeling units, some other methods performed multilingual learning by other ways. For example, Tan et al. \cite{tanxu2019-iclr} proposed to train separate models for each language at first and then perform knowledge distillation from each language-specific model to the multilingual model for multilingual translation. Wang et al. \cite{sokolov-nmt-G2P} presented a Grapheme-to-Phoneme (G2P) model which share the same encoder and decoder across multiple languages by utilizing a combination of universal symbol inventories of Latin-like alphabets and cross-linguistically shared feature representations. 
Inspired by these related methods, we make an exploration to the synergized learning of multilingual lip reading, by introducing phonemes as modeling units, and also a novel synchronous bidirectional learning framework to solve the multilingual lip reading problem, which has not been touched before.
\vspace{-0.4cm}
\section{The Proposed SBL Framework}
\vspace{-0.2cm}
\begin{figure}[ht]
	\setlength{\abovecaptionskip}{0cm}
	\setlength{\belowcaptionskip}{-0.2cm} 
	\centering
	\includegraphics[width=0.92\textwidth]{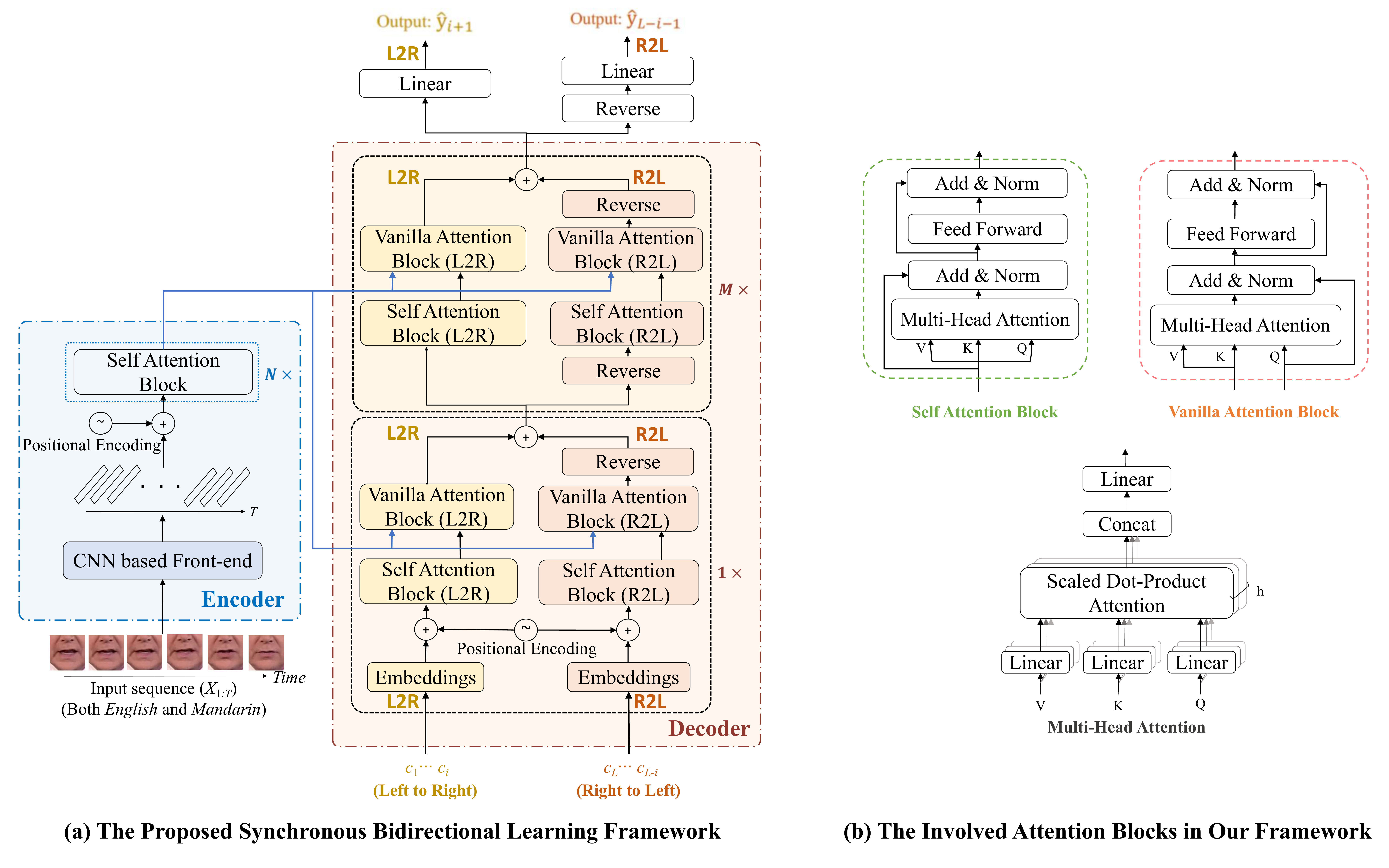}
	\caption{The whole framework of our model. The model takes the lip image sequence $\left(X_{1:T}\right)$ as input and outputs a sequence of phonemes $\bm{\hat{y}}_{1:L}$. During the inference process, the decoder employs the left-to-right (L2R) phoneme context $\bm{c}_{1:i}^{L2R} = [c_{1}, c_{2}, \dots, c_{i}]$ to predict $\hat{y}_{i+1}$, and the right-to-left (R2L) phoneme context $\bm{c}_{1:i}^{R2L}$ = [$c_{L}$, $c_{L-1}$, \dots, $c_{L-i}$] to predict $\hat{y}_{L-i-1}$. }\label{our_whole_model}
	\vspace{-0.4cm}
\end{figure}

 We build our model based on the Transformer architecture, as shown in Figure \ref{our_whole_model}. The whole model can be divided into two main parts: the visual encoder and the synchronous bidirectional decoder, which are shown as blue and yellow parts respectively in Figure \ref{our_whole_model}.(a). 
The visual encoder is responsible for encoding the input image sequence to a preliminary sequential representation of the sequence.
Then the synchronous bidirectional decoder is followed to take the outputs of the encoder as inputs and predicts both the left-to-right and the right-to-left output sequence simultaneously in the training process. 
By learning from the bidirectional context including both the previous and future time steps, the model could be able to learn the composition rules of each language.
%
% The idea of bidirectional learning has been proven effective in machine translation, such as \cite{ZhouZZ19}, \cite{ZhangZZZ20}. In this paper, we proposed a new synchronous bidirectional architecture for lip reading by combing the context from history and future to decode well in each time step.

\vspace{-0.4cm}
\subsection{\bf The Visual Encoder} 
\vspace{-0.1cm}
As shown in Figure \ref{our_whole_model}.(a), the visual encoder mainly consists of two modules, the CNN based front-end and $N$ stacked self-attention blocks. The CNN based front-end is used to capture the short-term spatial-temporal patterns in the image sequence, and $N$ stacked self-attention blocks are used to weight the patterns at different time steps in the visual sequence to obtain the final representation of the encoder.

Specifically, we denote the input image sequence as $X=(\text{x}_1, \text{x}_2, ..., \text{x}_T)$, where $T$ is the number of frames in the sequence. We use $H$ and $W$ to denote the height and width of the frames respectively. The image sequence is input to a 3D-convolutional layer firstly, followed by a max-pooling layer.
The spatial dimension is reduced to a quarter of the input size, while the temporal dimension is kept the same as the input. That is, the dimension of the output would be $T\times H/4\times W/4$. 
Then a ResNet-18 \cite{hekaiming_resnet_16} module is introduced to output a 512-\textit{d} vector at each time step, which would be added with their corresponding positional encodings and then used as the input of the subsequent self-attention blocks.  
The final output of the last self-attention block is taken as the final representation of the input sequence. We denote the output as $E_\theta(X)$, where $\theta$ represents the parameter of the encoder, and $E_\theta(X)$ is composed by $T$ 512-\textit{d} vectors.

The structure of each self-attention block is the same as \cite{vaswani2017attention}. As shown in Figure \ref{our_whole_model}.(b), the output of each self-attention block can be obtained as:
$$ 
\begin{array}
 {c}Q'=Q W_{Q}, K'=K W_{K}, V'=V W_{V},\\[1mm]
 {H(Q, K, V)=\textit{softmax}\left(\frac{Q' K'^{T}}{\sqrt{d_{k}}}\right) V'}, \\[1mm]
 {MH(Q, K, V) =\text{Concat}\left(H_{1}, \ldots, H_{\mathrm{h}}\right) W_{H}}.
\end{array}
\eqno{(1)}
$$
Where $H$ and $MH$ means the output of a single head attention block and a multi-head attention block respectively, $Q, K$ and $V$ equal to each other corresponding to the input of the block. Each head $H_j (j=1,..,h)$ would have its own learnable parameters $W_{Q}, W_K$ and $W_V$. $W_H$ is another learnable parameter to combine all the outputs from all the heads. In this paper, we employ $h=8$ and $N=6$ in the encoder.
The dimension $d_k$ of both the query matrix Q and key matrix $K$, and the dimension $d_v$ of the value matrix $V$, are all set to $64$.
\vspace{-0.3cm}
\subsection{The Synchronous Bidirectional Decoder}
\vspace{-0.1cm}
Given the representation $E_\theta (X)$ of each input sequence $X$, the synchronous bidirectional (SB) decoder is introduced to predict each phoneme $c_i$ at each output's time step $i$ ($i=1,2,\dots,L$). As shown in Figure \ref{our_whole_model}.(a), the decoder part is composed of several stacked synchronous bidirectional learning (SBL) blocks. Each block would combine the context from both the previous and future time steps to generate its output to the next SBL block. We use the context of the left-to-right (L2R) and the right-to-left (R2L) directions in the label sequence to express the previous and future context. 

As shown in Figure \ref{our_whole_model}.(a), each SBL block contains two branches: the L2R branch and the R2L branch. Each branch consists of a self-attention block and a vanilla-attention block. 
The self-attention block would perform a weighted sum of its input at different time steps, where the weights are obtained from the input by itself, as shown in Eq.(1) where $Q,K,V$ are all equal to the input. The vanilla attention block is similar to the self-attention block, and also output a weighted sum of its input ( which is corresponding to the output of the previous self-attention block) at different time steps. But the weights are generated according to the output of the encoder, as shown in Figure \ref{our_whole_model} where $K$ and $V$ are equal to the output of the encoder and $Q$ is the output of the previous self-attention block.

To effectively unify the L2R and the R2L branches, some differences exist between the first SBL block and the subsequent SBL blocks.
Specifically, we assume the ground truth labels of each sequence as $\bm{y}=(y_1, y_2, ..., y_L)$, where each sequence is padded to the same length $L$. The architecture can be described as follows.
\begin{itemize}\setlength{\itemsep}{0cm} 
    \item \textbf{For the first SBL block}, a sequence of phonemes before the current time step together with their corresponding positional encodings is used as the input. For example, when predicting the target unit at time step $i+1 (i=0,...,L-1)$, the input to the first L2R and R2L branch are $\bm{c}_{1:i}^{L2R}=(c_1, c_2,..., c_{i})$ and $\bm{c}_{1:i}^{R2L}=(c_{L}, c_{L-1},...,c_{L-i})$ respectively, as shown in Figure \ref{our_whole_model}.(a). Each $c_{i}$ (or $c_{L-i}$) is equal to the corresponding prediction result $\hat{y}_i$ (or $\hat{y}_{L-i}$) in the inference process. For training, we introduce probabilistic teacher forcing, where $c_{i}$ (or $c_{L-i}$) is equal to the ground truth unit $y_{i}$ (or $y_{L-i}$)with a probability $\gamma$, and to the previous prediction result $\hat{y}_{i}$ (or $\hat{y}_{L-i}$) with a probability 1-$\gamma$.%\hl{i $\rightarrow$ i-1; L-i $\rightarrow$ L-i+1}
    \item \textbf{For the SBL blocks after the first one}, the input from the previous output would be reversed at first to generate the R2L branch's input. 
    \item \textbf{For all the SBL blocks}, the output of R2L would be reversed at first to perform an element-wise summation with the output of L2R branch. Then the summation is used as the output of the corresponding SBL block.
    \vspace{-0.1cm}
\end{itemize}

Finally, two fully connected layers are introduced to project the output of the two branches of the last SBL block to the unified phoneme space respectively. 

To make the learning process more targeted and effective for each specific language, we also introduce an extra task to predict the language type of the input by adding an extra indicator label $F$ to the ground-truth sequence: $\bm{y} \rightarrow \{F, \bm{y}\}$. With the prediction task, the model can be guided to learn in a more targeted and effective manner for different languages.
\vspace{-0.8cm}
\subsection{Learning Process}
\vspace{-0.1cm}
Given the above pipeline, the model is learned by minimizing the cross-entropy loss at each time step. Specifically, we use $\hat{y_{i}}^{(L2R)}$ and $\hat{y_{i}}^{(R2L)}$ to denote the prediction results of the L2R and R2L branch at time step $i$ respectively. Then the model would be optimized to minimize $L_{total}$ as follows:
$$
L_{L2R}=-\sum_{i=1}^{L} {p(\hat{y_{i}}^{(L2R)}) \log p(\hat{y_{i}}^{(L2R)})}, 
~~~~~~L_{R2L}=-\sum_{i=1}^{L} {p(\hat{y}_{i}^{(R2L)})\log p(\hat{y}_{i}^{(R2L)})} \eqno{(2)}
$$
$$
L_{total}=\lambda_1 L_{L2R}+ \lambda_2 L_{R2L} \eqno{(3)}
$$
where $p(\hat{y_{i}}^{(L2R)})=p({\hat{y}_{i}^{(L2R)}|c_{1},c_{2},\dots,c_{i-1}})$, $p(\hat{y}_{L-i}^{(R2L)})=p({\hat{y}_{L-i}^{(R2L)}|c_{L},c_{L-1},\dots,c_{L-i+1}})$. $\lambda_1$ and $\lambda_2$ are used to balance the learning of the two branches, and both of them are set to be 0.5 in our experiments.

For the test process, we introduce the entropy of the prediction results of each branch to measure the quality of the corresponding branch. A smaller entropy of the prediction results indicates stronger confidence of the prediction. In the ideal case, the prediction is like a one-hot vector. 
We define $H(\hat{y}_{i}^{(L2R)})$ and $H(\hat{y}_{i}^{(R2L)}))$ as the entropy of the prediction distribution of the L2R and R2L branch at time step $i$ respectively. The combination result is denoted as \emph{\textbf{C-Bi}}, where $C$ means combining. It is achieved by:
$$ 
\textbf{C-Bi}_{i}=\left\{\begin{array}{ll} \hat{y}_{i}^{(L2R)} & {\text { if } H(\hat{y}_{i}^{(L2R)})<H(\hat{y}_{i}^{(R2L)})}. \\ {\hat{y}_{i}^{(R2L)}} & {\text { if } H(\hat{y}_{i}^{(L2R)})>H(\hat{y}_{i}^{(R2L)})}.\end{array}\right.
\eqno{(4)}
$$

In the setting where an extra language indicator flag $F$ is introduced, the above combination operation is performed only when the predictions of the language identity from both the two branches are the same. If the judgements are different, then we directly adopt the result from the branch which has a smaller entropy over the language identity judgement.

%%%%%以上已修改
\vspace{-0.3cm}
\section{Experiments}
Limited by the existing available large-scale lip reading datasets, we evaluate the proposed SBL framework with two languages, the English dataset LRW, and the Mandarin dataset LRW-1000. 

\textbf{English Lip Reading Dataset: LRW} \cite{chung2016lip}, released in 2016, is the first large scale English word-level lip-reading datasets, which includes 500 English words. There are 1000 training samples in each word class. All the videos are collected from BBC TV broadcasts, resulting in various types of speaking conditions in the wild. It has become a popular and influential benchmark for the evaluation of many existing lip reading methods.
% Most current methods perform word-level tasks using classification-based methods.

\textbf{Mandarin Lip Reading Dataset: LRW-1000} \cite{yang2019lrw-1000}, released in 2019, is a challenging and naturally distributed large scale benchmark for Mandarin word-level lip-reading. There are 1000 Mandarin words and phrases, and more than 700 thousand samples in total. The length and frequency of the words are all naturally distributed without extra limitations, forcing the model to be easily adaptive to the practical case where some words indeed appear more frequently than others. 

In our experiments, the split manner of training and test set is the same as divided by each dataset itself when we train mono-lingual lip reading models. When we train multilingual lip reading models, the training and test data is composed by the union of the training and test set of each language respectively. But the metric value is computed for each language separately to perform a comparison with other methods in a similar setting. 

%\hl{see the notation.Add illustrations about trn, tst..} % 增加一小段说明如何划分的训练测试数据。（单语言训练时，沿用数据集本身的训练测试划分；多语言训练时，各数据集的训练数据拿来训练，而测试集各自保持不变，作为各语言的识别性能测试）

\vspace{-0.25cm}
\subsection{\bf Implementation Details}
\vspace{-0.1cm}
We crop the mouth regions of each frame on LRW with a fixed bounding box of 112 by 112.%\hl{ xxx see the notation}. %←用大概1-2两句具体描述crop方式，读者在读到你的paper时直接就知道怎么复现而不用再去查找和下载别的paper
The images in LRW-1000 are already cropped well and we use them directly without other pre-processing. All the images are converted to grayscale, resized to 112 $\times$ 112, and then randomly cropped to 88 $\times$ 88. Each word in both the English dataset LRW and the Mandarin dataset LRW-1000 is converted to a sequence of phonemes, which would be used as the target label sequence. %Two examples of each language are shown in Figure \ref{illustration_our_idea}(c).
In our paper, we use 40, 48, and 56 phonemes for only English, only Chinese and the union of English and Chinese respectively.%, as shown in Table \ref{phonemes_set}.

In the training phase, the Adam \cite{adam14} optimizer is employed with default parameters. The learning rate would be changed automatically in the training process according to the number of training steps. %\hl{see the notation: }% 可否用一句话说明策略，尽量避免看你的论文需要反复查找和下载别的论文的情况 % the schedule strategy as \cite{vaswani2017attention}.
To speed up the training speed and ensure the generalization performance of the model, we set the teacher forcing rate $\gamma$ as 0.5.
The implementation is based on PyTorch. Dropout with probability 0.5 is applied to each layer in our model. 

To perform a convenient comparison with other methods, we also adopt the word-level accuracy (\emph{Acc.}) to measure our performance. For our model, \emph{{Acc.}}=1 - \emph{{WER}} where \emph{{WER}} is computed by comparing the predicted and ground truth phoneme sequence. We denote \emph{{PER}} as the phoneme error rate.  
\vspace{-0.35cm}
\subsection{Ablation Study of the Proposed SBL Framework}
\vspace{-0.1cm}
In this section, we try to answer two questions based on our model with a thorough comparison and analysis.
(1) Is it possible to perform multilingual lip reading, after all, mono-lingual lip reading itself has already been a very challenging task? (2) Would the proposed SBL framework be effective for the synergy of multilingual lip reading? How much improvement could it bring to the recognition of each specific language?

\vspace{0.2cm}
\noindent\textbf{I. For Question-1:} We answer it from the following comparison.
\vspace{-0.1cm}
\begin{itemize}\setlength{\itemsep}{0cm} 
    \item \textbf{I-A. TM (Baseline):} For the baseline, we use the visual encoder shown in Figure \ref{our_whole_model}.(a), with the decoder as the traditional Transformer \cite{vaswani2017attention}. Two different models are trained on the English and Mandarin lip reading datasets respectively, which is in the same way as traditional work. %For LRW, we use the English phonemes set as the modeling units. And for LRW-1000, we use the Chinese phonemes set as the modeling units. Both the two phonemes set are shown in Table \ref{phonemes_set}.
    \item \textbf{I-B. TM-ML:} Using the same architecture as I-A, but the model is trained with a new mixed data by combining LRW and LRW-1000 together, where \textbf{ML} refers to the introduction of training on different languages simultaneously. %. And we use the multilingual phonemes set as our modeling units, which is shown in Table \ref{phonemes_set}. We use 
    \item \textbf{I-C. TM-ML-Flag:} Based on I-B, we add an extra indicator flag to introduce an extra task of predicting the language type. Here, we use \textbf{Flag} to denote the introduction of this task. 
    \vspace{-0.2cm}
\end{itemize}

The results, in Table \ref{different_phonemes}, show the recognition performance of the model for both common phonemes and unique phonemes between the two languages, English and Mandarin (LRW and LRW-1000). We could find that the multilingual training can improve the ability of recognizing not only the common phonemes shared between these languages, but also the unique phonemes belonging to each specific language. At the same time, we find that the prediction performance has not much relation with the phoneme's position in the word. The multilingual setting would increase both the quantity and the diversity of each phoneme shared among different languages. At the same time, knowledge sharing and transfer among different languages can also improve the learning ability of the model. So the learning for the multilingual target could bring many benefits to the recognition.

The results are shown in Table \ref{baseline_BL_PER} and Table \ref{baseline_BL_BD}, which report both the phoneme error rate (\emph{\textbf{PER}}) and the accuracy (\emph{\textbf{Acc.}}=1-\emph{\textbf{WER}}) in different settings, where ``EN/CN'' and ``EN+CN'' mean that the corresponding model is trained with a single language or both lnaguages (EN: LRW and CN: LRW-1000).
According to Table \ref{baseline_BL_PER} and Table \ref{baseline_BL_BD}, we could find that there is a significant improvement when using mixed multilingual data for training. This shows that the joint learning of different languages could help improve the model's capacity and performance for phonemes, leading to enhance the model's performance for each individual language. This conclusion is consistent with the results in the related ASR and NLP domain \cite{dalmia2018sequence-based, graves2006connectionist, zhou2018multilingual, li2020towards}.

As can be seen from Table \ref{baseline_BL_PER} and Table \ref{baseline_BL_BD}, there is a further improvement when we further introduce an extra language type prediction task to the learning process. It suggests that an explicit introduction of the task to predict language type could help the model learn the rules of different languages more effectively.
\vspace{-0.2cm}

\begin{table}[H]
    \setlength{\abovecaptionskip}{-0.2cm}
    \small
    \begin{center}
        \begin{tabular}{|c|c|c|c|c|c|}    
            \hline
             \multirow{2}{*}{Method}& \multirow{2}{*}{Languages}& \multicolumn{2}{|c|}{EN\_LRW (\emph{\textbf{PER.}} $\downarrow$ )} & \multicolumn{2}{|c|}{CN\_LRW-1000 (\emph{\textbf{PER.}} $\downarrow$ )} \\
             \cline{3-6}
             ~&~&\textbf{CPs}&\textbf{UPs}&\textbf{CPs}&\textbf{UPs}\\
            \hline
            \hline
            \textbf{TM(Baseline)}&  EN/CN& 16.12\%& 17.58\%& 48.03\%& 48.90\% \\
            %\hline
            \textbf{TM-ML}&  EN+CN& 13.85\%& 14.97\%& 46.81\% & 47.55\% \\
            %\hline
            \textbf{TM-ML-Flag}&  EN+CN& 13.76\%& 14.45\%& 46.68\%& 47.03\% \\
            %\hline
            \textbf{TM-ML-BD}&  EN+CN& 13.05\% & 13.77\% & 41.91\% & 43.52\% \\
            \textbf{TM-ML-BD-Flag}&  EN+CN& 12.88\% &13.50\% & 41.79\% & 42.44\% \\
            %\hline
            \hline
        \end{tabular}
    \end{center}
    \caption{Evaluation of the effects of multilingual synergized learning for common and unique phonemes prediction. \textbf{CPs}:Common Phonemes, \textbf{UPs}:Unique Phonemes. $\downarrow$ means that the lower the value is, the better the performance is.} \label{different_phonemes}
    \vspace{-0.4cm}
\end{table}
%\hl{see the notation: Add an extra experimental reuslt.}%这里增加一小段和一个表格总结，用以说明①无论音素位置在哪里②无论是否是共享音素，音素的准确率都有提升；给出共享音素的准确率变化、非共享音素的准确率变化，说明无论是哪种都有提升；
\begin{table}[H]
    \setlength{\abovecaptionskip}{-0.2cm}
    \small
    \begin{center}
        \begin{tabular}{|c|c|c|c|c|c|c|c|}    
            \hline
             \multirow{2}{*}{Method}& \multirow{2}{*}{Languages}& \multicolumn{3}{|c|}{EN\_LRW (\emph{\textbf{PER.}} $\downarrow$ )} & \multicolumn{3}{|c|}{CN\_LRW-1000 (\emph{\textbf{PER.}} $\downarrow$ )} \\
             \cline{3-8}
             ~&~&L2R&R2L&\emph{\textbf{C-Bi}}&L2R&\multicolumn{1}{|c|}{R2L}&\emph{\textbf{C-Bi}}\\
            \hline
            \hline
            \textbf{TM(Baseline)}&  EN/CN& \--&\--& 16.98\%& \--&\-- &48.42\% \\
            %\hline
            \textbf{TM-ML}&  EN+CN& \--&\--& 14.53\%& \--&\-- &47.21\% \\
            %\hline
            \textbf{TM-ML-Flag}&  EN+CN& \--&\--& 14.12\%& \--&\-- &46.83\% \\
            %\hline
            \textbf{TM-ML-BD}&  EN+CN& 13.50$\%$&13.66$\%$&13.37$\%$ & 43.82$\%$&42.71$\%$&42.35$\%$ \\
            \textbf{TM-ML-BD-Flag}&  EN+CN& 13.39$\%$&13.53$\%$&13.19$\%$ & 43.11$\%$&42.20$\%$&42.03$\%$ \\
            %\hline
            \hline
        \end{tabular}
    \end{center}
    \caption{Evaluation of the effects of multilingual synergized learning for phonemes prediction. \textbf{TM}:Transformer, \textbf{ML}:Multi-lingual, \textbf{BD}:Bi-directional, $\downarrow$ means that the lower the value is, the better the performance is.} \label{baseline_BL_PER}
    \vspace{-0.4cm}
\end{table}

\noindent\textbf{II. For Question-2:} We perform comparison and analysis from two aspects. Firstly, we evaluate our idea that the composition rules of each language can be learned more easily by using bi-directional context and so could provide help for multilingual lip reading. Then we compare with the proposed SBL framework to verify its effectiveness. For this target, we performed the following comparison at first.
\vspace{-0.2cm}
\begin{itemize}\setlength{\itemsep}{0cm} 
    \item \textbf{II-A. TM-ML-BD:} Based on the setting of I-B, we introduce an extra decoder module which is targeted to make predictions in a right-to-left direction. We use \textbf{BD} to denote that bi-directional information is used in the learning process.
    \vspace{-0.1cm}
    \item \textbf{II-B. TM-ML-BD-Flag:} Based on the model II-A, an extra prediction task of language type as I-C is introduced in this setting. 
    \vspace{-0.1cm}
\end{itemize}
\vspace{-0.1cm}

The results are shown in  Table \ref{baseline_BL_PER} and Table \ref{baseline_BL_BD}. We can see that there is an obvious improvement when the bi-directional context is introduced. The accuracy increased from 81.03\% and 44.58\% to 84.12\% and 52.61\% on LRW and LRW-1000 respectively. This improvement verifies the effectiveness of our idea that the rules of each language could be learned by learning to infer the target phoneme given its bidirectional context. When we introduce the extra task of predicting language type, the performance is further improved.  
\vspace{-0.2cm}
\begin{table}[H]
    \setlength{\abovecaptionskip}{-0.2cm}
    \small
    \begin{center}
        \begin{tabular}{|c|c|c|c|c|c|c|c|}    
            \hline
             \multirow{2}{*}{Method}& \multirow{2}{*}{Languages}& \multicolumn{3}{|c|}{EN\_LRW (\emph{\textbf{Acc.}} $\uparrow$ )} & \multicolumn{3}{|c|}{CN\_LRW-1000 (\emph{\textbf{Acc.}} $\uparrow$ )} \\
             \cline{3-8}
             ~&~&L2R&R2L&\emph{\textbf{C-Bi}}&L2R&\multicolumn{1}{|c|}{R2L}&\emph{\textbf{C-Bi}}\\
            \hline
            \hline
            \textbf{TM(Baseline)}&  EN/CN& \--&\--& 76.22\%& \--&\-- &41.83\% \\
            %\hline
            \textbf{TM-ML}&  EN+CN& \--&\--& 81.03\%& \--&\-- &44.58\% \\
            %\hline
            \textbf{TM-ML-Flag}&  EN+CN& \--&\--& 82.17\%& \--&\-- &45.24\% \\
            %\hline
            \textbf{TM-ML-BD}&  EN+CN& 83.56$\%$&82.78$\%$&84.12$\%$ & 49.33$\%$&51.48$\%$&52.61$\%$ \\
            \textbf{TM-ML-BD-Flag}&  EN+CN& 84.04$\%$&83.26$\%$&84.63$\%$ & 50.35$\%$&52.67$\%$&53.29$\%$ \\
            %\hline
            \hline
        \end{tabular}
    \end{center}
    \caption{Evaluation of the baseline methods for multilingual lip reading. $\uparrow$ means that the higher the value is, the better the performance is.} \label{baseline_BL_BD}
    \vspace{-0.4cm}
\end{table}
\vspace{-0.05cm}

\begin{table}[H]
    \setlength{\abovecaptionskip}{0.0cm}
    \small
    \begin{center}
        \begin{tabular}{|c|c|c|c|c|c|c|c|}    
            \hline
             \multirow{2}{*}{Method}& \multirow{2}{*}{Languages}& \multicolumn{3}{|c|}{EN\_LRW (\emph{\textbf{Acc.}} $\uparrow$ )} & \multicolumn{3}{|c|}{CN\_LRW-1000 (\emph{\textbf{Acc.}} $\uparrow$ )} \\
             \cline{3-8}
             ~&~&L2R&R2L&\emph{\textbf{C-Bi}}&L2R&\multicolumn{1}{|c|}{R2L}&\emph{\textbf{C-Bi}}\\
            \hline
            \hline            
            \textbf{TM-ML-BD}&  EN+CN& 83.56$\%$&82.78$\%$&84.12$\%$ & 49.33$\%$&51.48$\%$&52.61$\%$ \\
            \textbf{TM-ML-BD-Flag}&  EN+CN& 84.04$\%$&83.26$\%$&84.63$\%$ & 50.35$\%$&52.67$\%$&53.29$\%$ \\
            \textbf{SBL-First}& EN+CN& 84.97$\%$ & 83.46$\%$&85.26$\%$ &51.79$\%$&53.82$\%$&54.35$\%$ \\
            %\hline
            \textbf{SBL-All}&   EN+CN& 86.21$\%$ &85.04$\%$ &86.78$\%$&52.78$\%$&55.63$\%$&56.12$\%$  \\
            %\hline
            \textbf{SBL-All-Flag}&   EN+CN& 86.88$\%$ &85.64$\%$ &87.32$\%$&53.41$\%$&56.29$\%$&56.85$\%$  \\
            \hline
        \end{tabular}
    \end{center}
    \caption{The SBL results for exploring multilingual lip reading. $\uparrow$ means that the higher the value is, the better the performance is.} \label{SBL_ablation}
    \vspace{-0.4cm}
\end{table}
\vspace{-0.1cm}
Based on the above evaluation, we make a further comparison of the above bidirectional models with our proposed SBL, which unifies the two-directional context together in a single block, instead of two separate single-directional modules. For this target, we perform the following experiments. 
\vspace{-0.1cm} 
\begin{itemize}\setlength{\itemsep}{0cm} 
    % \vspace{-0.1cm}
    \item \textbf{II-C. SBL-First:} In this setting, we only introduce the first SBL module to the decoder, but keep the subsequent blocks in the decoder as the traditional blocks in the vanilla Transformer \cite{vaswani2017attention}. 
    \vspace{-0.1cm}
    \item \textbf{II-D. SBL-All:} In this setting, the architecture is totally the same as shown in Figure \ref{our_whole_model}.(a), where each block in the decoder is designed to combine the bidirectional context together. 
    \vspace{-0.1cm}
    \item \textbf{II-E. SBL-Flag:} This setting is almost the same as II-D, except that an extra task of predicting language type is introduced to the learning process.
    \vspace{-0.1cm}
\end{itemize}
% \vspace{-0.1cm}
\vspace{-0.1cm}
The results are shown in Table \ref{SBL_ablation}. As we can see, it is much better even we introduce the SBL block only at the first layer. It achieves the performance of 85.26\% and 54.35\% on LRW and LRW-1000 respectively. This result has already outperformed \textbf{TM-ML-BD-Flag} which introduce two separate uni-directional decoder branches and the extra prediction task of language type. When we introduce the SBL block through the whole decoder with the extra language-type prediction task, \textbf{SBL-All-Flag} outperforms the others by a large margin on both the two datasets. 
\subsection{Comparison with the State of the Art}
\vspace{-0.1cm}
In this part, we perform a comparison with other related state-of-the-art lip reading methods, including both seq2seq based decoding methods and classification based methods, as shown in Table \ref{comparison_with_sota}. In the table, \cite{petridis2018end}, \cite{StafylakisT17}, \cite{xiao2020deformation}, \cite{zhao2020mutual}, \cite{zhang2020can} are based on sequential classification structures. And \cite{zhang2019spatio-temporal}, \cite{luo2020pseudo-convolutional} are based on sequential decoding structures. We can find that our SBL framework outperforms the state-of-the-art performance by a large margin, especially on LRW-1000. One noteworthy result is that \cite{zhang2019spatio-temporal} achieved an accuracy of $83.7$\% on the English benchmark LRW after pre-training on two extra large-scale English lip reading datasets, LRS2-BBC and LRS3-TED. But their result is worse than ours which use only an extra Mandarin dataset LRW-1000 which has a smaller scale than LRS2-BBC and LRS3-TED. This result could provide another support to the benefits of multilingual training.

%\hl{see the notation.Add description about [20].}% 增加对[20]的描述，指出他们先用了与LRW同语言的LRS数据集做预训练，但依然比我们的低很多。而LRS比LRW-1000大很多。这个对比可以说明multi-lingual lipreading的可行性。并简要分析为什么多语言训练比它好，是只因为数据多了吗还是怎样（参考rebuttal的回答）。
%\hl{Add illustration about multiple languags.see the notation.} %增加评审中的说明，本文的方法虽然只在两种语言上进行的测试，但对于多语言也适用，为什么？
%About the difference of improvement, we think that it's possibly due to the data format.
% compared to other most works. 
\begin{table}[H]
    \setlength{\abovecaptionskip}{0.00cm}
    %\setlength{\abovecaptionskip}{0cm}
    %\vspace{-0.3cm}  %调整图片与上文的垂直距离
    \centering
    \begin{tabular}{|c|c|c|c|}    
        \hline
        Work & Method & LRW & LRW-1000\\
        \hline
        \hline
        %\cite{chung2016lip}  & 61.10$\%$ & - \\
        %\cite{ChungZ18} & 71.50$\%$ & - \\
        %\cite{chung2017lip}&76.20$\%$&-  \\
        \cite{StafylakisT17}-2017 &Classifying & 83.00$\%$& - \\
        \cite{petridis2018end}-2018&Classifying &82.00$\%$&- \\
        \cite{zhang2019spatio-temporal}-2019&Decoding &83.70$\%$&-\\
        % \cite{yang2019lrw-1000}-2019&-&38.19$\%$ \\
        % \cite{wang2019multi-grained}-2019&83.34$\%$&36.91$\%$\\
        \cite{luo2020pseudo-convolutional}-2020&Decoding &83.50$\%$&38.70$\%$\\
        \cite{zhao2020mutual}-2020&Classifying &84.41$\%$&38.79$\%$\\
        \cite{zhang2020can}-2020&Classifying &85.02$\%$&45.24$\%$\\
        \cite{xiao2020deformation}-2020&Classifying &84.13$\%$&41.93$\%$\\
        %\emph{\textbf{Ours(Bb-multi)}}& 84.12$\%$&52.61$\%$\\
        \textbf{Ours (SBL-First)}&Decoding & \textbf{85.26}$\%$&\textbf{54.35}$\%$\\
        \textbf{Ours (SBL-All)}&Decoding& \textbf{86.78}$\%$&\textbf{56.12}$\%$\\
        \textbf{Ours (SBL-All-Flag)}& Decoding &\textbf{87.32}$\%$&\textbf{56.85}$\%$\\
        \hline
    \end{tabular}    
    \vspace{0.2cm}
    \caption{Comparison with other related methods.} \label{comparison_with_sota}
\end{table}
\vspace{-0.8cm}
\section{Conclusion}
\vspace{-0.2cm}
Inspired by the related multilingual study in the field of automatic speech recognition and NLP, we try to explore the possibility of multilingual synergized lip reading with large scale datasets for the first time. The phonemes are introduced as the modeling units to bridge different languages. And a new synchronous bidirectional learning manner is introduced to unify the two-directional context together in each block, to enhance the learning of each language. Both the proposed model and the learning process are not related to some specific properties of some single language, so it can be directly employed to three or more languages. Limited by the available large-scale lip reading datasets, we perform a thorough evaluation and analysis on the English and Mandarin datasets. Our work achieves new state-of-the-art performance on both the two challenging benchmarks, LRW (English) and LRW-1000 (Mandarin). 
\vspace{-0.4cm}
\section{Acknowledgments}
\vspace{-0.2cm}
This work is partially supported by National Key R\&D Program of China (No. 2017YFA0700800) and National Natural Science Foundation of China (No. 61702486, 61876171).

% %\bibliographystyle{bmvc}
\bibliography{egbib}
\end{document}